\pdfoutput=1
\documentclass[journal]{IEEEtran}

\usepackage{blindtext}
\usepackage{graphicx}
\usepackage{booktabs}
\usepackage{subcaption}
\usepackage{multirow} 
\usepackage{bigdelim} 
\usepackage{xspace}
\usepackage{bbm}
\usepackage{array}
\usepackage{multicol}
\usepackage{lipsum}
\usepackage{xspace}
\usepackage{amsmath}
\usepackage{amsfonts}
\usepackage{amssymb}
\usepackage{hyperref}
\usepackage{colortbl}
\definecolor{verylightgray}{gray}{0.9}
\usepackage{algorithm}
\usepackage{algpseudocode}
\usepackage{resizegather} 

\newcommand{\bigO}[1]{\ensuremath{\mathop{}\mathopen{}O\mathopen{}\left(#1\right)}}
\newcommand{\TV}{\ensuremath{\mathrm{TV}}}

\newcommand{\prox}{\ensuremath{\mathrm{prox}}}

\renewcommand{\matrix}[1]{\ensuremath{\mathbf{#1}}}
\renewcommand{\vector}[1]{\ensuremath{\mathbf{#1}}}
\newcommand{\tran}[1]{\ensuremath{{#1}^{\scriptscriptstyle\top}}}
\newcommand{\T}{\scriptscriptstyle\top}

\newcommand{\norm}[2]{\ensuremath{{\left\| #2 \right\|}_{#1}}}

\newcommand{\Frob}[1]{\ensuremath{{\left\| #1 \right\|}_{F}}}

\newcommand{\opt}[1]{\ensuremath{{#1}^{*}}}

\def\nSamples{\ensuremath{N}}
\def\nDim{\ensuremath{P}}
\def\nComp{\ensuremath{K}}
\def\loneCoeff{\ensuremath{\lambda_{1}}}
\def\ltwoCoeff{\ensuremath{\lambda_{2}}}

\def\structCoeff{\ensuremath{\lambda}}
\def\X{\matrix{X}}
\def\U{\matrix{U}} 
\def\u{\vector{u}}
\def\V{\matrix{V}} 
\def\v{\vector{v}}
\def\bbeta{\vector{v}}
\def\y{\frac{\X^{\T}\u}{n\ltwoCoeff}}
\def\bsigma{\vector{\sigma}}
\def\Id{\matrix{I}}
\def\D{\matrix{D}} 

\def\A{\matrix{A}}

\def\Loss{\mathcal{L}}

\def\setG{\mathcal{G}}
\def\setK{\mathcal{K}}
\def\setR{\ensuremath{\mathbb{R}}}

\def\balpha{\boldsymbol{\alpha}}

\newcommand{\ie}{\textit{i.e.}\xspace}

\newcommand{\eg}{\textit{e.g.}\xspace}

\renewcommand{\eqref}[1]{Equation~\ref{#1}}
\newcommand{\figref}[1]{Figure~\ref{#1}}
\newcommand{\tableref}[1]{Table~\ref{#1}}
\newcommand{\secref}[1]{Section~\ref{#1}}

\begin{document}

\title{Structured Sparse Principal Components Analysis with the TV-Elastic Net penalty}
%
%
%

\author{
Amicie de Pierrefeu,
Tommy L\"ofstedt,
Fouad Hadj-Selem,
Mathieu Dubois,
Philippe Ciuciu,
Vincent Frouin,
Edouard Duchesnay
\thanks{A. de Pierrefeu, E. Duchesnay, P. Ciuciu, M. Dubois and V. Frouin are with NeuroSpin, CEA, Paris-Saclay, Gif-sur-Yvette - France}
\thanks{F. Hadj-Selem is with the Energy Transition Institute: VeDeCoM - France.}
\thanks{T. L\"ofstedt is with Department of Radiation Sciences, Ume\aa{} University, Ume\aa{} - Sweden.}
}

%
%

\markboth{IEEE Transactions on Medical Imaging}
{De Pierrefeu \MakeLowercase{\textit{et al.}}: Bare Demo of IEEEtran.cls for Journals}
%



\maketitle

\begin{abstract}
Principal component analysis (PCA) is an exploratory tool widely used in data analysis to uncover dominant patterns of variability within a population. Despite its ability to represent a data set in a low-dimensional space, PCA's interpretability remains limited. Indeed, the components produced by PCA are often noisy or exhibit no visually meaningful patterns. Furthermore, their high density may also impede interpretation, unless arbitrary thresholding is used.  However, in neuroimaging, it is essential to uncover clinically interpretable phenotypic  markers  that would account  for  the  main  variability  in the brain images of a population. Recently, some alternatives to the standard PCA approach, such as Sparse PCA, have been proposed, their aim being to limit the density of the components. Nonetheless, sparsity alone does not entirely solve the interpretability problem in neuroimaging, since it may yield scattered and unstable components. We hypothesized that the incorporation of prior information regarding the structure of the data may lead to improved relevance and interpretability of brain patterns. We therefore present a simple extension of the popular PCA framework  that adds structured sparsity penalties on the loading vectors in order to identify the few stable regions in the brain images accounting for most of the variability. Such structured sparsity can be obtained by combining \eg, $\ell_1$ and total variation (TV) penalties, where the TV regularization encodes higher order information about the structure of the data. This paper presents the structured sparse PCA (denoted SPCA-TV) optimization framework and its resolution. We demonstrate SPCA-TV's efficiency and versatility on three different data sets. It can be applied to any kind of structured data, such as \eg, $N$-dimensional array images or meshes of cortical surfaces. The gains of SPCA-TV over unstructured approaches are significant, since SPCA-TV reveals the variability within a data set in the form of intelligible brain patterns that are easy to interpret, and are more stable across different samples.
\end{abstract}

\begin{IEEEkeywords}
MRI, unsupervised machine learning, PCA, total variation.
\end{IEEEkeywords}

\graphicspath{ {figures/} }

\IEEEpeerreviewmaketitle

\section{Introduction}
Principal components analysis (PCA) is an unsupervised statistical procedure 
whose aim is to capture dominant patterns of variability in order to provide an 
optimal representation of a data set in a lower-dimensional space defined by the 
principal components (PCs).
Given a data set $\X  \in \setR^{\nSamples \times \nDim}$ of $\nSamples$ samples 
and $\nDim$ centered variables, PCA aims to find the most accurate rank-$K$ approximation
of the data:
\begin{align} \label{pb:SVD_pb}
\min_{\U, \D, \V} &~ \Frob{\X - \U \D \V^T}^2,\\
\mathrm{s.t.} &~ \U^T\U = \Id, \V^T\V =\Id, d_1 \geq \cdots \geq d_K > 0 \nonumber
\end{align}
where $\Frob{.}$ is the Frobenius norm of a matrix, $\V = [\v_1, \cdots, \v_K] 
\in \setR^{\nDim\times\nComp}$ are the $\nComp$ loading vectors (right singular 
vectors) that define the new coordinate system where the original features 
are uncorrelated,
$\D$ is the diagonal matrix of the $\nComp$ singular values, and
$\U = [\u_1, \cdots, \u_K] \in \setR^{\nSamples\times\nComp}$  are the $\nComp$ 
projections of the original samples in the new coordinate system (called 
principal components (PCs) or left singular vector).
Using $\nComp=\text{rank}(\X)$  components leads to the singular value decomposition
(SVD).
A vast majority of neuroimaging problems involve high-dimensional feature spaces 
($\approx 10^5$ features \ie voxels, mesh vertices) with a relatively limited sample size  ($\approx 
10^2$ participants). With such ``large $\nDim$, small $\nSamples$'' problems, the 
SVD formulation, based on the data matrix, is much more efficient than an 
eigenvalue decomposition of the large $\nDim \times  \nDim$ covariance matrix.

In a neuroimaging context, our goal is to discover the phenotypic markers accounting for the main variability in a population's brain images. For example, when considering structural images of patients that will convert to Alzheimer disease (AD), we are interested in revealing the brain patterns of atrophy explaining the 
variability in this population. This provides indications of possible 
stratification of the cohort into homogeneous sub-groups that may be clinically similar but with a different pattern of atrophy. This could suggest different sub-types of patients with AD or some other etiologies such as dementia with Lewy bodies. Clustering methods might be natural approaches to address such situations, however, they can not reveal subtle differences 
that go beyond a global and trivial pattern of atrophy. Such patterns are
usually captured by the first component of PCA which, after being removed, offers the possibility to identify spatial patterns on the subsequent components.
However, PCA provides dense loading vectors (patterns), that cannot be used to identify brain markers without arbitrary thresholding.

Recently, some alternatives propose to add sparsity in this matrix 
factorization problem. The sparse dictionary learning framework proposed by 
\cite{mairal_sDL_2010} provides a sparse coding (rows of $\U$) of samples through 
a sparse linear combination of dense basis elements (columns of $\V$). However, 
the identification of biomarkers requires a sparse dictionary (columns of $\V$).
This is precisely the objective of Sparse PCA (SPCA) proposed in 
\cite{jolliffe_modified_2003, zou_sparse_2006, daspremont_direct_2007, 
witten_penalized_2009, journee_generalized_2010} which adds a sparsity-inducing 
penalty on the columns of $\V$.

However, sparse PCA is limited by the fact that it ignores the inherent spatial correlation in the data. It leads to 
scattered patterns that are difficult to interpret. Furthermore, constraining 
only the number of features included in the PCs might not always be fully relevant 
since most data sets are expected to have a spatial structure. For 
instance,
MRI data is naturally encoded on a grid; some voxels are neighbors, 
while others
are not. We hypothesize that brain patterns are organized in 
regions rather
than scattered across the brain. Recent studies tried to overcome this limitation by encoding prior information 
concerning the spatial structure of the data (see \cite{Jenatton09, Guo2015, Wang2015}).
However, they used methods that are difficult to plug 
into the optimization scheme (\eg, spline smoothing, wavelet smoothing) and 
incorporated prior information that sometimes may be difficult to define.
In data classification problems, when extracting structured and sparse 
predictive maps, the goals are largely aligned with those of PCA. Some classification 
studies have revealed stable and interpretable results by adding a total 
variation (TV) penalty to the sparsity constraint (see \cite{Dubois2014}).
TV is widely used as a tool in image denoising and reconstruction. It accounts 
for the spatial structure of images.

We propose to extend the PCA of \eqref{pb:SVD_pb} by adding structured
sparsity-inducing penalties (TV and $\ell_1$) to the minimization problem:
\begin{align} \label{pb:pca-l1-l2-struct}
    \min_{\U, \D, \V}
    &\, \frac{1}{\nSamples}\Frob{\X - \U \D \tran{\V}}^2  \nonumber\\
    &\, + \sum_{k=1}^{\nComp} \bigg\{ \ltwoCoeff \norm{2}{\v_k}^2 + \loneCoeff \norm{1}{\v_k} + \structCoeff\sum_{g \in \setG} \| \A_g \bbeta\|_2 \bigg\},\\
    \mathrm{s.~t.} & ~\norm{2}{\u_k}^2 = 1, \forall k=1, \cdots, \nComp, \nonumber
\end{align}
where $\loneCoeff$, $\ltwoCoeff$ and $\structCoeff$ are hyper-parameters
controlling the relative strength of each penalty.
We further propose a generic optimization framework that can combine any differentiable convex
(penalized) loss function with: (i) penalties whose proximal operator is known (here $\norm{1}{\cdot}$) and (ii) a
large range of complex, non-smooth convex structured penalties that can be formulated
as a $\norm{2,1}{\cdot}$-norm defined over a set of groups $\setG$. Such group-penalties cover \eg, total variation and overlapping group lasso.
%
This new problem aims at finding a linear combination of original variables that 
points in directions explaining as much variance as possible in data while
enforcing sparsity and structure (smoothness for TV) of the loadings.

To achieve this, it is necessary to sacrifice some of the explained variance as well as the 
orthogonality of both the loading and the principal components. 
Most existing SPCA algorithms \cite{zou_sparse_2006, daspremont_direct_2007, 
witten_penalized_2009, journee_generalized_2010}, do not impose orthogonal 
loading directions either.
While we forced the components to have unit norm for visualization purposes, we do
not, in this formulation, enforce $\norm{2}{\v_k}=1$. Instead, the value of 
$\norm{2}{\v}$ is controlled by the  hyper-parameter $\ltwoCoeff$. This penalty 
on the loading, together with the unit norm constraint on the component, prevents 
us from obtaining trivial solutions.
The optional $\frac{1}{\nSamples}$ factor acts on and conveniently normalizes the loss to account for the number of samples in order to simplify the 
settings of the hyper-parameters: $\loneCoeff, \ltwoCoeff, \structCoeff$.

This paper presents an extension of the popular PCA framework by adding structured
sparsity-inducing penalties on the loading vectors in order to identify the few stable regions in the brain images accounting for most of the variability.
The addition of a prior that reflects the data's structure within the learning
process gives the paper a scope that goes beyond Sparse PCA. To our knowledge, very
few  papers (\cite{Abraham13, Guo2015,Jenatton09,Wang2015}) addressed the use of structural constraint in PCA. Only one study, recently used the total variation prior (\cite{Abraham13}), in a context of multi-subject dictionary learning, based on a different optimization scheme (\cite{beck09}).

Section \ref{sec:method} presents our main contribution: a simple optimization algorithm
that combines well known methods (deflation scheme and alternate minimization)
with an original continuation algorithm based on Nesterov's smoothing technique.
Our proposed algorithm has the ability to include the TV penalty, but many other non-smooth penalties, such as \eg overlapping group lasso, could also be used.
This versatile mathematical framework is an essential feature in neuroimaging.
Indeed, it enables a straightforward application to all kinds of data with known 
structure such as N-dimensional images (of voxels) or meshes of (cortical) 
surfaces.
Section \ref{sec:experiments} demonstrates the relevance of structured sparsity
on both simulated and experimental data, for structural and fMRI acquisitions.
Structured sparsity achieved a higher reconstruction accuracy and more stable
solutions than classical elastic net or Sparse PCA.





\section{Method} \label{sec:method}

A common approach to solve the PCA problem, see~\cite{daspremont_direct_2007, journee_generalized_2010, witten_penalized_2009}), is to compute a rank-1 approximation of the data matrix, and then repeat this on the deflated matrix~\cite{mackey_deflation_2009}, where the 
influence of the PCs are successively extracted and discarded. We first detail the notations of the problem for a single component estimation (\secref{sec:single_comp}), and its solution using an alternating minimization pipeline (\secref{sec:algo:power}).
Then, we develop the TV regularization framework (\secref{sec:tv} and \secref{sec:Nesterov}). Lastly, we will discuss the algorithm used to solve the minimization problem and its ability to converge toward stable pairs of components/loading vectors (\secref{sec:conesta}) and (\secref{sec:algo}).

\subsection{Single component computation} \label{sec:single_comp}

Given a a pair of loading/component vectors, $\u \in \setR^{\nSamples}, \v \in \setR^{\nDim}$,
the best rank-1 approximation of the problem given in \eqref{pb:pca-l1-l2-struct} is equivalent \cite{witten_penalized_2009} to:
\begin{align}\label{pb:pca-l1-l2-struct_single}
    \min_{\u, \v} f \equiv &~
    \overbrace{
      \underbrace{-\frac{1}{N}\tran{\u} \X \v + \ltwoCoeff \norm{2}{\v}^2 \vphantom{\sum_g G}}_{l(\bbeta)}
    }^{\text{smooth}}
    + \overbrace{\vphantom{\frac{1}{n}}
       \loneCoeff\underbrace{\norm{1}{\bbeta}\vphantom{\sum_g G}}_{h(\bbeta)}
    + \structCoeff\underbrace{\sum_{g \in \setG} \| \A_g \bbeta\|_2}_{s(\bbeta)}
    }^{\text{non-smooth}}\\
    \mathrm{s.~t.} &~ \norm{2}{\u}^2 \leq 1, \nonumber
\end{align}
where $l(\bbeta)$ is the penalized smooth (\ie differentiable) loss, $h(\bbeta)$ is a sparsity-inducing penalty whose
proximal operator is known and $s(\bbeta)$ is a complex penalty on the structure of the input variables with an unknown proximal operator.
This problem is convex in $\u$ and in $\v$ but not in $(\u, \v)$.

\subsection{Alternating minimization of the bi-convex problem} \label{sec:algo:power}

The objective function to minimize is bi-convex~\cite{Boyd}. The most common approach
to solve a bi-convex optimization problem (which does not guarantee global optimality
of the solution) is to alternatively update $\u$ and $\v$ by fixing one of them at
the time and solving the corresponding convex optimization problem on the other
parameter vector.

On the one hand, when $\v$ is fixed, the problem to solve is
\begin{align} \label{pb:pca-l1-l2-struct_single_u}
    \min_{\u \in \setR^{\nSamples}} &~ -\frac{1}{\nSamples}\tran{\u} \X \v \\
    \mathrm{s.~t.} &~ \norm{2}{\u}^2 \leq 1, \nonumber
\end{align}
with the associated explicit solution
\begin{equation} \label{eq:opt_u_fixed_v_pb}
    \opt{\u}(\v) = \frac{\X\v}{\norm{2}{\X\v}}.
\end{equation}
On the other hand, solving the equation with respect to
$\v$ with a fixed $\u$ presents a higher level of difficulty that will be discussed in Section~\ref{sec:conesta}.

\subsection{Reformulating TV as a linear operator} \label{sec:tv}
Before discussing the minimization with respect to $\v$, we provide details on the
encoding of the spatial structure within the $s(\v)$ penalty.
It is essential to note that the algorithm is independent of the spatial structure of the data.
All the structural information is encoded in a linear operator, $\A$, that is computed outside of the algorithm.
Thus the algorithm has the ability to address various structured data and, most
importantly, other penalties than just the TV penalty. The algorithm requires the setting of two parameters: (i) the linear 
operator $\A$; (ii) a projection function detailed in \eqref{eq:projection_l2}.
This section presents the formulation and the design of $\A$ in the specific case of a TV penalty on the loading vector $\v$ measured on a 3-dimensional image or a mesh of the cortical surface.

\subsubsection{3D image} The brain mask is used to establish a mapping $g(i, j, k)$ between
the coordinates $(i, j, k)$ in the 3D grid, and an index $g \in [\![1; \nDim]\!]$ 
in the collapsed image. We extract the spatial neighborhood of $g$, of size $\leq 4$, corresponding to voxel $g$ and its 3 neighboring voxels, within the mask, in the $i, j$ and $k$ directions.
By definition, we have
\begin{equation} \label{eq:tv_def}
     \TV(\bbeta) \equiv  \sum_{g = 1}^{\nDim} \big\| \nabla \left( \bbeta_{g(i, j, k)} \right) \big\|_2.
\end{equation}
The first order approximation of the spatial gradient $\nabla(\bbeta_{g(i, j, k)})$ is computed by applying the linear operator  $\A_g^{'} \in \mathbb{R}^{3 \times 4}$ 
to the loading vector $\bbeta_g$ in the spatial neighborhood of $g$, \ie
\begin{equation} \label{eq:grad_Ag}
\nabla\left( \bbeta_{g(i, j, k)} \right) = \underbrace{\left[
    \begin{array}{cccc}
        -1 & 1 & 0 & 0 \\ 
        -1 & 0 & 1 & 0 \\
        -1 & 0 & 0 & 1
    \end{array}\right]}_{\A_g^{'}}
    \underbrace{\left[
    \begin{array}{l}
        v_{g(i,j,k)}\\
        v_{g(i+1,j,k)}\\
        v_{g(i,j+1,k)}\\
        v_{g(i,j,k+1)}
    \end{array}\right]}_{\bbeta_g},
\end{equation}
where $v_{g(i,j,k)}$ is the loading coefficient at index $g$ in the collapsed image corresponding to
voxel $(i, j, k)$ in the 3D image.
Then $\A_g^{'}$ is extended, using zeros, to a large but very sparse matrix $\A_g \in \mathbb{R}^{3 \times \nDim}$ in order to be directly applied on the full vector $\bbeta$.
If some neighbors lie outside the mask, the corresponding rows in $\A_g$ are removed.
Noticing that for TV there is one group per voxel in the mask ($\setG = [\![1; \nDim]\!]$), 
we can reformulate TV from \eqref{eq:tv_def} using a general expression:
\begin{equation} \label{s_asl12}
    \TV(\bbeta) =  \sum_{g \in \setG} \| \A_g \bbeta \|_2.
\end{equation}
Finally, with a vertical concatenation of all the $\A_g$ matrices, we obtain the 
full linear operator $\A \in \mathbb{R}^{3\nDim \times \nDim}$ that will be used 
in Section~\ref{sec:conesta}.

\subsubsection{Mesh of cortical surface}
The linear operator $\A_g^{'}$ used to compute a first order approximation of the
spatial gradient can be obtained by examining the neighboring vertices of each vertex
$g$. With common triangle-tessellated surfaces, the neighborhood size is $\leq 7$
(including $g$). In this setting, we have $\A_g^{'} \in \mathbb{R}^{3 \times 7}$,
which can be extended and concatenated to obtain the full linear operator $\A$.

\subsection{Nesterov's smoothing of the structured penalty} \label{sec:Nesterov}
We consider the convex non-smooth minimization of \eqref{pb:pca-l1-l2-struct_single}
with respect to $\v$, where thus $\u$ is fixed. This problem includes a general
structured penalty, $s(\cdot)$, that covers the specific case of TV. 
A widely used approach when dealing with non-smooth problems is to use methods based on the proximal operator of the penalties. For the $\ell_1$ penalty alone, the proximal operator is analytically known and efficient iterative algorithms such as ISTA and FISTA are available (see~\cite{Beck2009}).
However, since the proximal operator of the TV$+\ell_1$ penalty is not known to have an analytical solution, standard implementation of those algorithms is not suitable. In order to overcome this barrier, we used Nesterov's smoothing technique~\cite{Nesterov2005a}. It consists of approximating the non-smooth penalties for which the proximal operator is unknown (\eg, TV) with a smooth function (of which the gradient is known). Non-smooth penalties with known proximal operators (\eg, $\ell_1$) are not affected. Hence, as described in~\cite{Chen2012}, it allows to use an exact accelerated proximal gradient algorithm. Thus we can solve the PCA problem penalized by TV and elastic net, where an exact $\ell_1$ penalty is used.

Using the dual norm of the $\ell_2$-norm (which happens to be the $\ell_2$-norm again),
\eqref{s_asl12} can be reformulated as
\begin{equation} \label{eq:s_asl1_dual}
    s(\bbeta) =  \sum_{g \in \setG} \| \A_g \bbeta \|_2 = \sum_{g \in \setG} \max_{\|\balpha_g\|_2\leq 1} \tran{\balpha}_g \A_g \bbeta,
\end{equation}
where $\balpha_g \in \setK_g = \{\balpha_g \in \mathbb{R}^{3}: \|\balpha_g\|_2\leq 1\}$ is a vector of auxiliary variables, in the $\ell_2$ unit ball, associated with $\A_g \v$.
As with $\A \in \mathbb{R}^{3\nDim \times \nDim}$ which is the vertical concatenation of all the $\A_g$,
we concatenate all the $\balpha_{g}$ to form the $\balpha \in \setK=\{[\balpha_1^{\T}, \ldots, \balpha_{\nDim}^{\T}]^{\T}: \balpha_g \in \setK_g\} \in \mathbb{R}^{3\nDim}$.
$\setK$ is the Cartesian product of 3D unit balls in Euclidean space and, therefore, a compact convex set. \eqref{eq:s_asl1_dual} can further be written as
\begin{equation}\label{eq:s_dual}
   s(\v) = \max_{\balpha \in K} \balpha^{\T} \A \v.
\end{equation}
Given this formulation of $s(\v)$, we can apply Nesterov's smoothing. For a given
smoothing parameter, $\mu > 0$, the $s(\v)$ function is approximated by the smooth
function
\begin{equation}\label{eq:s_dual_mu}
   s_{\mu}(\v) = \max_{\balpha \in K} \left\lbrace \balpha^{\T} \A \v - \frac{\mu}{2}\|\balpha\|_2^2 \right\rbrace,
\end{equation}
for which $\lim_{\mu \rightarrow 0} s_{\mu}(\v) = s(\v)$. Nesterov~\cite{Nesterov2005a} demonstrates this convergence using the inequality in \eqref{eq:nesterov_inequality}.
The value of $\balpha_\mu^{*}(\v) = [\balpha_{\mu, 1}^{*\T}, \ldots, \balpha_{\mu, g}^{*\T}, \ldots ,\balpha_{\mu, \nDim}^{*\T}]^{\T}$ that maximizes \eqref{eq:s_dual_mu} is the concatenation of projections of vectors $A_g\v \in \mathbb{R}^{3}$ to the $\ell_2$ ball ($\setK_g$): $\balpha^*_{\mu, g}(\v) = \mathrm{proj}_{\setK_g}\left( \frac{A_g\v}{\mu} \right)$, 
where
\begin{equation}\label{eq:projection_l2}
 \mathrm{proj}_{\setK_g}(\vector{x}) = \begin{cases}
                            \vector{x} & \text{~if~} \|\vector{x}\|_2 \leq 1\\ 
                            \frac{\vector{x}}{\|\vector{x}\|_2} & \text{~otherwise.} \\
                        \end{cases}.
\end{equation}

The function $s_\mu$, \ie the Nesterov's smooth transform of $s$, is convex and differentiable. Its gradient given by~\cite{Nesterov2005a}
\begin{equation} \label{eq:s_mu:grad}
    \nabla(s_{\mu})(\v) = \A^{\T}\balpha^*_{\mu}(\v),
\end{equation}
which is Lipschitz-continuous with constant
\begin{equation} \label{eq:s_mu:Lipschitz}
L\big(\nabla(s_{\mu})\big) = \frac{\|\A\|_2^2}{\mu},
\end{equation}
where \(\|\A\|_2\) is the matrix spectral norm of $\A$.
Moreover, Nesterov~\cite{Nesterov2005a} provides the following inequality relating
$s_\mu$ and $s$
\begin{equation} \label{eq:nesterov_inequality}
    s_\mu(\bbeta)\leq s(\bbeta)\leq s_\mu(\bbeta) + \mu M,\quad \forall \bbeta \in \mathbb{R}^p,
\end{equation}
where $M = \max_{\balpha \in \setK} \frac{\|\balpha\|_2^2}{2}=\frac{\nDim}{2}$.
    
Thus, a new (smoothed) optimization problem, closely related to
\eqref{pb:pca-l1-l2-struct_single} (with fixed $\u$),
arises from this regularization as
\begin{gather}\label{pb:pca-l1-l2-struct_mu_single}
    \min_{\v}
      \overbrace{
      \underbrace{-\frac{1}{n}\tran{\u} \X \v\!
      +\ltwoCoeff \norm{2}{\v}^2}_{l(\v)}
      + \structCoeff \underbrace{\left\lbrace{\balpha_\mu^{*}(\v)}^{\T} \A \v - \frac{\mu}{2}\|\balpha^*\|_2^2\right\rbrace}_{s_\mu(\v)}
      }^{\text{smooth}}
      + \loneCoeff%
        \hspace{-0.7em}%
        \overbrace{\underbrace{\vphantom{\Big(}\norm{1}{\v}}_{h(\v)}
      }^{\text{non-smooth}}.
\end{gather}

%
Since we are now able to explicitly compute the gradient of the smooth part $\nabla(l+\structCoeff s_\mu)$ (\eqref{eq:g+s_mu:grad}), its Lipschitz constant (\eqref{eq:g+s_mu:Lipschitz}) and also the proximal operator of the non-smooth part, we have all the ingredients necessary to solve this minimization function using an accelerated proximal gradient methods~\cite{Beck2009}.
Given a starting point $\bbeta^0$ and a smoothing parameters $\mu$, FISTA (Algorithm \ref{algo:fista}) minimizes the smoothed problem and reaches a prescribed precision $\varepsilon_\mu$.

However, in order to control the convergence of the algorithm (presented in \secref{sec:duality_gap}), we introduce the Fenchel dual function and the
corresponding dual gap of the objective function. The Fenchel duality requires
the loss to be strongly convex, which is why we further reformulate \eqref{pb:pca-l1-l2-struct_mu_single} slightly: All penalty terms are divided by $\ltwoCoeff$ and by using the following equivalent formulation for the loss, we
obtain the minimization problem
\begin{gather}\label{pb:pca-l1-l2-struct_mu_single_reformulate}
    \min_{\bbeta} f_{\mu}  \equiv
      \lefteqn{\overbrace{\phantom{\frac{1}{2}\norm{2}{\bbeta - \y}^2+\frac{1}{2}\norm{2}{\bbeta}^2}}^{l(\bbeta)}} 
       \underbrace{\frac{1}{2}\norm{2}{\bbeta - \y}^2}_{\Loss(\bbeta)}
      +\underbrace{\frac{1}{2}\norm{2}{\bbeta}^2
      + \frac{\structCoeff}{\ltwoCoeff} \overbrace{\vphantom{\norm{2}{\y}^2}
      \left\lbrace{\balpha_\mu^{*}(\bbeta)}^{\T} \A \bbeta - \frac{\mu}{2}\|\balpha^*\|_2^2\right\rbrace }^{s_\mu(\bbeta)}
      +\frac{\loneCoeff}{\ltwoCoeff} \overbrace{\vphantom{\norm{2}{\y}^2}
      \norm{1}{\bbeta}}^{h(\bbeta)}}_{\psi_{\mu}(\bbeta)}.
\end{gather}
This new formulation of the smoothed objective function (noted $f_{\mu}$) preserves the decomposition of $f_{\mu}$ into a sum
of a smooth term $l+ \frac{\structCoeff}{\ltwoCoeff} s_\mu$ and a non-smooth term $h$. 
Such decomposition is required for the application of FISTA with Nesterov's smoothing.
Moreover, this formulation provides a decomposition of $f_{\mu}$ into a sum of a smooth loss $\Loss$ and a penalty term $\psi_{\mu}$ required for the calculation of the gap presented in \secref{sec:duality_gap})

We provide all the required quantities to minimize \eqref{pb:pca-l1-l2-struct_mu_single_reformulate} using Algorithm~\ref{algo:fista}.
Using \eqref{eq:s_mu:grad} we compute the gradient of the smooth part as
\begin{align}\label{eq:g+s_mu:grad}
\nabla\left(l+ \frac{\structCoeff}{\ltwoCoeff} s_\mu\right) &=  \nabla(l) + \frac{\structCoeff}{\ltwoCoeff} \nabla(s_{\mu}) \nonumber\\
&= (2\bbeta - \y) + \frac{\structCoeff}{\ltwoCoeff} \A^{\T}\balpha^*_{\mu}(\bbeta^k),
\end{align}
and its Lipschitz constant (using \eqref{eq:s_mu:Lipschitz})
\begin{align}\label{eq:g+s_mu:Lipschitz}
L\left(\nabla\left(l+ \frac{\structCoeff}{\ltwoCoeff} s_\mu\right)\right) &= 2 + \frac{\structCoeff}{\ltwoCoeff} \frac{\|\A\|_2^2}{\mu}.
\end{align}

\begin{algorithm} 
  \caption{FISTA\big($\X^{\T}\u$, $\bbeta^0$, $\varepsilon_\mu$, $\mu$, $\A$, $\lambda$, $L(\nabla(g))$\big)}\label{algo:fista}
  \begin{algorithmic}[1]
      \State $\bbeta^1 = \bbeta^0$; $k=2$ 
      \State Compute the gradient of the smooth part $\nabla(g+\structCoeff s_\mu)$ (\eqref{eq:g+s_mu:grad}) and its Lipschitz constant $L_{\mu}$ (\eqref{eq:g+s_mu:Lipschitz}).
      \State Compute the size $t_\mu = L_{\mu}^{-1}$\label{algo:fista:step}
      \Repeat
	\State $\mathbf{z} = \bbeta^{k-1} + \frac{k-2}{k+1}\left(\bbeta^{k-1}-\bbeta^{k-2}\right)$
	\State $\bbeta^k = \prox_{h}\big(\mathbf{z} - t_\mu\nabla (g+\structCoeff s_\mu) (\mathbf{z})\big)$ 
      \Until{$\textsc{Gap}_{\mu}(\bbeta^{k}) \leq \varepsilon_\mu$}\label{algo:fista:stop}
      \State \textbf{return} $\bbeta^{k}$
  \end{algorithmic}
\end{algorithm}

\subsection{Minimization of the loading vectors with CONESTA} \label{sec:conesta}

The step size $t_\mu$ computed in Line~\ref{algo:fista:step} of Algorithm \ref{algo:fista}, depends on the smoothing parameter $\mu$ (see \eqref{eq:g+s_mu:Lipschitz}).
Hence, there is a trade-off between speed and precision. Indeed, high precision, with a small $\mu$, will lead to a slow convergence (small $t_\mu$). Conversely, poor precision (large $\mu$) will lead to rapid convergence (large $t_\mu$). Thus we propose a continuation approach (Algorithm \ref{algo:conesta}) which decreases the smoothing parameter with respect to the distance to the minimum. On the one hand, when we are far from $\bbeta^*$ (the minimum of \eqref{pb:pca-l1-l2-struct_mu_single_reformulate}), we can use a large $\mu$ to rapidly decrease the objective function. On the other hand,  when we are close to $\bbeta^*$, we need a small $\mu$ in order to obtain an accurate approximation of the original objective function.

\subsubsection{Duality gap} \label{sec:duality_gap}
The distance to the unknown $f(\bbeta^*)$ is estimated using the duality gap.
Duality formulations are often used to control the achieved precision level when
minimizing convex functions. They provide an estimation of the
error $f(\bbeta^k) - f(\bbeta^*)$, for any $\bbeta$, when the minimum is unknown.
The duality gap is the cornerstone of the CONESTA algorithm. Indeed, it is used three times:
\begin{enumerate}\renewcommand{\labelenumi}{\roman{enumi}}
    \item As the stopping criterion in the inner FISTA loop (Line \ref{algo:fista:stop} in Algorithm~\ref{algo:fista}). FISTA will stop as soon as the current precision is achieved using the current smoothing parameter, $\mu$.
      This prevents unnecessary convergence toward the approximated (smoothed) objective function.
    \item In the $i$th CONESTA iteration, as a way to estimate the current error $f(\bbeta^i) - f(\bbeta^*)$ (Line \ref{algo:conesta:eps} in Algorithm~\ref{algo:conesta}).
      The error is estimated using the gap of the smoothed problem $\textsc{Gap}_{\mu=\mu^i} (\bbeta^{i+1})$ which avoid unnecessary computation since it has already been computed during the last iteration of FISTA.
      The inequality in \eqref{eq:nesterov_inequality} is used to obtain the gap $\varepsilon^i$ to the original non-smoothed problem.
      The next desired precision $\varepsilon^{i+1}$ and the smoothing parameter, $\mu^{i+1}$, are derived from this value.
    \item Finally, as the global stopping criterion within CONESTA (Line \ref{algo:conesta:stop} in Algorithm~\ref{algo:conesta}).
      This will guarantee that the obtained approximation of the minimum, $\bbeta^{i}$, at convergence, satisfies $f(\bbeta^{i}) - f(\bbeta^*) < \varepsilon$.
\end{enumerate}

Based on \eqref{pb:pca-l1-l2-struct_mu_single_reformulate}, which decomposes the smoothed objective function as a sum of a strongly convex loss and the penalty,
$$
f_{\mu}(\bbeta) = \Loss (\bbeta) + \psi_{\mu}(\bbeta),
$$
we compute the duality gap that provides an upper bound estimation of the error to the optimum.
At any step $k$ of the algorithm, given the current primal $\bbeta^k$ and the dual $\bsigma(\bbeta^k)\equiv\nabla \Loss(\bbeta^k)$ variables~\cite{borwein2006convex}, we can compute the duality gap using the Fenchel duality rules~\cite{Mairal2010}:
\begin{equation}\label{gap}
    \textsc{Gap}(\bbeta^k) \equiv f_{\mu}(\bbeta^k) + \Loss^*\big(\bsigma(\bbeta^k)\big) + 
\psi_{\mu}^*\big(-\bsigma(\bbeta^k)\big),
\end{equation}
where $\Loss^*$ and $\psi_{\mu}^*$ are respectively the Fenchel conjugates of $\Loss$ and $\psi_{\mu}$. 
Denoting by $\bbeta^*$ the minimum of $f_{\mu}$ (solution of \eqref{pb:pca-l1-l2-struct_mu_single_reformulate}), the interest of the duality gap is that it provides an upper bound for the difference with the optimal value of the function. Moreover, it vanishes at the minimum:
\begin{equation} \label{eq:gap_prop}
    \begin{array}{rl}
        \textsc{Gap}(\bbeta^k) \geq f(\bbeta^k) - f(\bbeta^*) &\geq\;\, 0 \\
        \textsc{Gap}(\bbeta^*)                                &=\;\,    0.
    \end{array}
\end{equation}
The dual variable is
\begin{equation} \label{eq:dual_var}
\bsigma(\bbeta^k)\equiv\nabla \Loss(\bbeta^k)=\v - \frac{\X^{\T}\u}{n\ltwoCoeff},
\end{equation}
the Fenchel conjugate of the squared loss $\Loss(\bbeta^k)$ is
\begin{equation} \label{eq:l_star}
\Loss^*(\bsigma(\bbeta^k))=\frac{1}{2}\|\bsigma(\bbeta^k)\|_2^2+\bsigma(\bbeta^k)^{\T}\y.
\end{equation}
In \cite{hadj-selem_iterative_2016} the authors provide
the expression of the Fenchel conjugate of the penalty $\psi_{\mu}(\bbeta^k)$:
\begin{align} \label{eq:penalty_conj}
\psi_{\mu}^*(-\bsigma(\bbeta^k)) =& \frac{1}{2}\sum_{j=1}^{\nDim} \left(\bigg[\Big|-\bsigma(\bbeta^k)_j - \frac{\structCoeff}{\ltwoCoeff} \big(\A^{\T} \balpha_{\mu}^*(\bbeta^k)\big)_j\Big| - \frac{\loneCoeff}{\ltwoCoeff}\bigg]_+^2\right)\nonumber\\
                           &+ \frac{\structCoeff\mu}{2\ltwoCoeff}\big\|\balpha_{\mu}^*(\bbeta^k)\big\|_2^2
\end{align}
where $[\cdot]_+ = \max(0, \cdot)$.

The expression of the duality gap in \eqref{gap} provides an estimation of the distance to the minimum. This distance is geometrically decreased by a factor $\tau=0.5$ at the end of each continuation, and the decreased value defines the precision that should be reached by the next iteration (Line \ref{algo:conesta:update_eps} of Algorithm \ref{algo:conesta}).
Thus, the algorithm dynamically generates a sequence of decreasing prescribed precisions $\varepsilon^i$. Such a scheme ensures the convergence~\cite{hadj-selem_iterative_2016} towards a globally desired final precision, $\varepsilon$, which is the only parameter that the user needs to provide.

\subsubsection{Determining the optimal smoothing parameter} \label{sec:optimal_mu}
Given the current prescribed precision $\varepsilon^i$, we need to compute an optimal smoothing parameter $\mu_{opt}(\varepsilon^i)$ (Line~\ref{algo:conesta:update_mu} in Algorithm~\ref{algo:conesta}) that minimizes the number of FISTA iterations needed to achieve such precision when minimizing \eqref{pb:pca-l1-l2-struct_single} (with fixed $\u$) via \eqref{pb:pca-l1-l2-struct_mu_single_reformulate} (\ie, such that $f(\v^{(k)}) - f(\v^*) < \varepsilon^i$).

In~\cite{hadj-selem_iterative_2016}, the authors provide
the expression of this optimal smoothing parameter:
\begin{equation} \label{eq:mu_opt}
    \mu_{opt}(\varepsilon^i) = \frac{-\structCoeff M \|\A\|_2^2 + \sqrt{(\structCoeff M \|\A\|_2^2)^2 + M L(\nabla(l))\|\A\|_2^2\varepsilon^i}}{ M L(\nabla(l))},
\end{equation}
where $M=\nDim/{2}$ (\eqref{eq:nesterov_inequality}) and $L(\nabla(l))=2$ is the Lipschitz constant of the gradient of $l$ as defined in \eqref{pb:pca-l1-l2-struct_mu_single_reformulate}.

We call the resulting algorithm CONESTA (short for \textbf{CO}ntinuation with
\textbf{NE}sterov smoothing in a \textbf{S}hrinkage-\textbf{T}hresholding
\textbf{A}lgorithm). 
It is presented in detail, with convergence proofs in~\cite{hadj-selem_iterative_2016}.

Let $K$ be the total number of FISTA loops used in CONESTA, then we have experimentally verified that the convergence rate to the solution of \eqref{pb:pca-l1-l2-struct_mu_single} is $\bigO{1/K^2}$ (which is the optimal convergence rate for first-order methods). Also, the algorithm works even if some of the weights $\loneCoeff$ or $\structCoeff$ are zero, which thus allows us to solve \eg, the elastic net using CONESTA. Note that it has been rigorously proved that the continuation technique improves the convergence rate compared to the simple smoothing using a single value of $\mu$. Indeed, it has been demonstrated in \cite{teb2} (see also \cite{Chen2012}) that the convergence rate obtained with single value of $\mu$, even optimised, is $\bigO{1/K^2} + \bigO{1/K}$. However, it has recently been proved in \cite{hadj-selem_iterative_2016} that the CONESTA algorithm achieves a $\bigO{1/K}$ for general convex functions.

We note that CONESTA could easily be adapted to many other penalties.
For example, to add the group lasso (GL) constraint to our structure, we just have to design a specific linear operator $\A_{GL}$ and concatenate it to the actual linear operator $\A$.

\begin{algorithm}
  \caption{CONESTA\big($\X^{\T}\u$, $\varepsilon$\big)}\label{algo:conesta}
  \begin{algorithmic}[1]
      \State Initialize $\bbeta^0 \in \setR^{\nDim}$
      \State $\varepsilon^0  = \tau \cdot \textsc{Gap}_{\mu=10^{-8}}(\bbeta^0)$
      \State $\mu^0= \mu_{opt}\left(\varepsilon^0 \right)$
      \Repeat
	\State $\varepsilon_\mu^i =\varepsilon^i-\mu^i \gamma M$ \label{algo:conesta:eps_mu}
        \State $\bbeta^{i+1} =$ \textsc{Fista}($\X^{\T}\u$, $\bbeta^{i}$, $\varepsilon_\mu^i$, \ldots)
	\State $\varepsilon^i = \textsc{Gap}_{\mu=\mu^i} (\bbeta^{i+1}) + \mu^i \gamma M$ \label{algo:conesta:eps}
        \State $\varepsilon^{i+1} = \tau \cdot \varepsilon^i$\label{algo:conesta:update_eps}
        \State $\mu^{i+1} = \mu_{opt}\big(\varepsilon^{i+1}  \big)$ \label{algo:conesta:update_mu}
      \Until{$\varepsilon^i \leq \varepsilon$}\label{algo:conesta:stop}
      \State \textbf{return} $\bbeta^{i+1}$
  \end{algorithmic}
\end{algorithm}

\subsection{The algorithm for the SPCA-TV problem} \label{sec:algo}

The computation of a single component through SPCA-TV can be achieved by combining CONESTA and \eqref{eq:opt_u_fixed_v_pb} within an alternating minimization loop. Mackey~\cite{mackey_deflation_2009} demonstrated that further components can be efficiently obtained by incorporating this single-unit procedure in a deflation scheme as done in \eg~\cite{daspremont_direct_2007, journee_generalized_2010}. 
The stopping criterion is defined as
\begin{equation} \label{SPCATV:stopping_criterion}
    \textsc{StoppingCriterion} =  \frac{\Frob{\X^k - \u^{i+1} {\v^{i+1}}^{\T}} - \Frob{\X^k - \u^{i} {\v^{i}}^{\T}}}{\Frob{\X^k - \u^{i+1} {\v^{i+1}}^{\T}}}.
\end{equation}

All the presented building blocks were combined into Algorithm \ref{algo:spca-tv}
to solve the SPCA-TV problem.

\begin{algorithm}
  \caption{SPCA-TV($\X$, $\varepsilon$\big)}\label{algo:spca-tv}
  \begin{algorithmic}[1]
    \State $\X_{0} = \X$
    \ForAll{$k = 0, \ldots, K$} \Comment{Components}
	\State Initialize $\u^0 \in \setR^{\nSamples}$
	\Repeat \Comment{Alternating minimization}
	\State $\v^{i+1} = \textsc{CONESTA}(\X_{k}^{\T}\u^i, \varepsilon)$
	\State $\u^{i+1} =  \frac{\X_k\v^{i+1}}{\norm{2}{\X_k\v^{i+1}}}$
	\Until {$\textsc{StoppingCriterion} \leq \varepsilon$}\label{algo:Alternating:stop}
	\State $\v_{k+1} = \v^{i+1}$
	\State $\u_{k+1} = \u^{i+1}$
	\State $\X_{k+1} = \X_{k} - \u^{k+1} {\v^{k+1}}^{\T}$ \Comment{Deflation}
    \EndFor
    \State \textbf{return} $\U = [\u_1, \cdots, \u_K], \V = [\v_1, \cdots, \v_K]$
  \end{algorithmic}
  
\end{algorithm}

\section{Experiments} \label{sec:experiments}

We evaluated the performance of SPCA-TV using three experiments: One simulation study carried out on a synthetic data set and two on neuroimaging data sets.
In order to compare the performance of SPCA-TV with existing Sparse PCA models, we also included results obtained with Sparse PCA and ElasticNet PCA. We used the scikit-learn implementation \cite{scikit-learn} for the Sparse PCA while we used the Parsimony package (\url{https://github.com/neurospin/pylearn-parsimony}) for the ElasticNet PCA and the SPCA-TV methods. The number of parameters to set for each method is different: ElasticNet PCA requires the setting of the $\ell_1$ and the $\ell_2$ penalties weights. Meanwhile, SPCA-TV requires the settings of an additional parameter; the spatial constraint penalty $\lambda$. We operated a re-parametrization of these penalty weights in ratios. A global parameter controls the weight attributed to the whole penalty term, including the TV and the $\ell_1$ regularization. Individual constraints are expressed in terms of ratios: the $\ell_1$ ratio: $\lambda_1/(\lambda_1+\lambda_2+\lambda)$ and the TV ratio : $\lambda/(\lambda_1+\lambda_2+\lambda)$. For each of these methods, we tested several ranges of parameters, and retained the combination of parameters leading to the lowest reconstruction error obtained on an independent test set. However, in order to ensure that the components extracted have a minimum amount of sparsity, we also included a criteria controlling sparsity: At least half of the features of the second and third components have to be zero.

For both real neuroimaging experiments, performance was evaluated through a resampling procedure using a 5-fold cross-validation (5CV). However, for the synthetic data set, performance was evaluated on 50 different purposely-generated data sets. In order to assess the reconstruction accuracy, we reported the mean Frobenius norm of the reconstruction error across the folds/data sets, on independent test data sets. The hypothesis we wanted to test was whether there was a substantial decrease in the reconstruction error when using SPCA-TV compared to when using Sparse PCA or ElasticNet PCA. It was tested through a related two samples $t$-test.

The TV penalty has a more important purpose than just to minimize the reconstruction error: the estimation of coherent and reproducible loadings. Indeed, clinicians expect that, if images from other patients with comparable clinical conditions had been used, the extracted loading vectors would have turned out to be similar. The stability of the loading vectors obtained across various training data sets (variation in the learning samples) was assessed through a similarity measure: the pairwise Dice index between loading vectors obtained with different folds/data sets \cite{Dice1945}. We tested whether pairwise Dice indices are significantly higher in SPCA-TV compared to the Sparse PCA and ElasticNet PCA methods. Testing this hypothesis is equivalent to testing the sign of the difference of pairwise Dice indices between methods. However, since the pairwise Dice indices are not independent from one another (the folds share many of their learning samples), the direct significance measures are biased. We therefore used permutation testing to estimate empirical $p$-values. The null hypothesis was tested by simulating samples from the null distribution. We generated $1\,000$ random permutations of the sign of the difference of pairwise Dice index between the PCA methods under comparisons, and then the statistics on the true data were compared with the ones obtained on the permuted ones to obtain empirical $p$-values.

\subsection{Simulation study}

\begin{figure}[!bh]
    \centering
    \includegraphics[width=0.4825\textwidth]{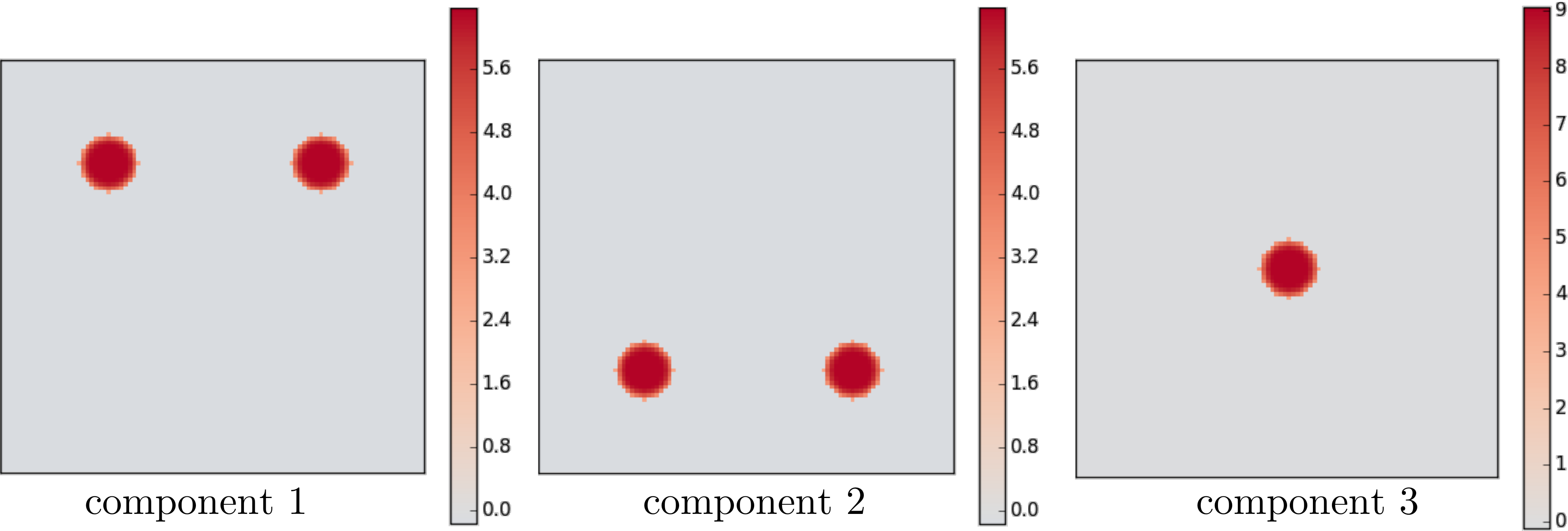}
    \caption{Loading vectors $V = [V^{1},V^{2},V^{3}] \in \mathbb{R}^{10\,000 \times 3}$ used to generate the images}
    \label{fig:latent_variables}
\end{figure}

We generated 50 sets of synthetic data, each composed of 500 images of size $100\times100$ pixels. Images are generated using the following noisy linear system : 

\begin{equation}
    u_1V^{1} + u_2V^{2} + u_3V^{3} +  \epsilon \in \mathbb{R}^{10\,000},
\end{equation}
where  $V = [V^{1},V^{2},V^{3}] \in \mathbb{R}^{10\,000 \times 3}$ are  sparse and structured loading vectors, illustrated in  \figref{fig:latent_variables}. The support of $V^{1}$ defines the two upper dots, the  support of $V^{2}$ defines the two lower dots, while $V^{3}$ 's support delineates the middle dot. The coefficients  $u = [u_{1},u_{2},u_{3}]$ that linearly combine the components of V are generated according to a centered Gaussian distribution. The elements of the noise vector $\epsilon $ are independent and identically distributed according to a centered Gaussian distribution with a 0.1 signal-to-noise ratio (SNR). This SNR was selected by a previous calibration pipeline, where we tested the efficiency of data reconstruction at multiple SNR ranges running from 0 to 0.5. We decided to work with a 0.1 SNR because it is located in the range of values where standard PCA starts being less efficient in the recovery process. 

We split the 500 artificial images into a test and a training set, with 250 images in each set and with the decomposition learned on the training set. The optimal number of components was selected according to the explained variance accounted for by each component. We decided to retain the information accounted for by the first three components, since the next PCs offers very little increase in the total explained variance. This choice seems reasonable in our attempt to recover the underlying support of the three loading vectors.

\begin{figure}[!bh]
    \centering
    \includegraphics[width=0.4825\textwidth]{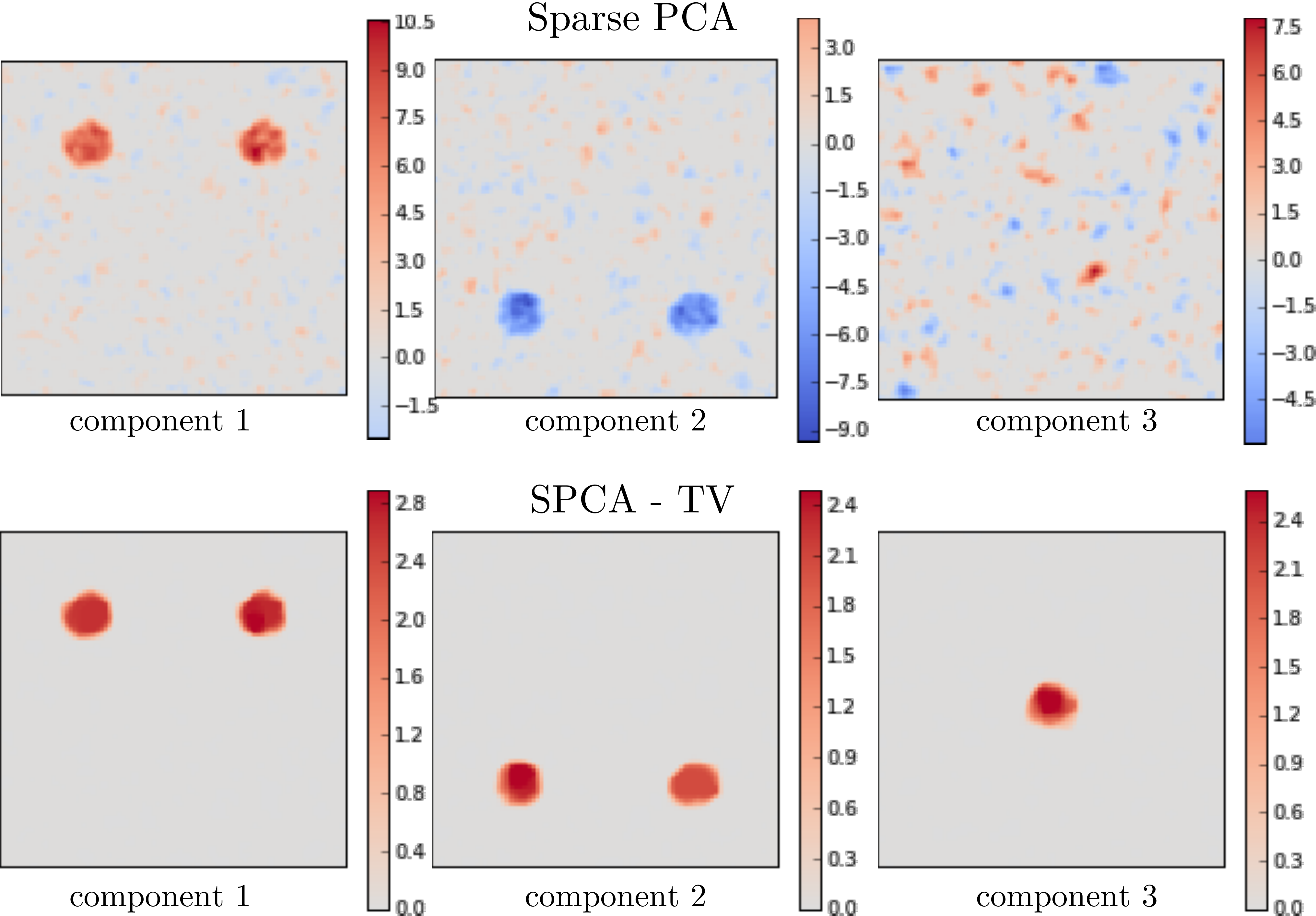}
    \caption{Loading vectors recovered from 250 images using Sparse PCA and SPCA-TV.}
    \label{fig:extracted_components}
\end{figure}

\figref{fig:extracted_components} represents the loading vectors extracted with one data set. We observe that Sparse PCA yields very scattered loading vectors. The loading vectors of SPCA-TV, on the other hand, are sparse; but also organized in clear regions. SPCA-TV provides loading vectors that closely match the ground truth. The reconstruction error is evaluated on the test sets (\tableref{tab:simulation_score}), with its value over the 50 data sets being significantly lower in SPCA-TV than in both Sparse PCA ($T=-46.1$, $p=4.8\cdot10^{-42}$) and ElasticNet PCA ($T=24.0$, $p=8.5\cdot10^{-29}$) methods.  A different way of quantifying the reconstruction accuracy for each method is to evaluate how closely the extracted loadings match the known ground truth of simulated data set. We computed the mean squared error (MSE) between the ground truth and the estimated loadings. The results are presented in \tableref{tab:simulation_score}. We note that the MSE is significantly lower with SPCA-TV than with both Sparse PCA  ($T=10.58$, $p=2.95\cdot10^{-14}$) ElasticNet PCA ($T=8.5$, $p=2.8\cdot10^{-11}$) .

Moreover, when evaluating the stability of the loading vectors across resampling, we found a higher statistically significant mean Dice index when using SPCA-TV compared to the other sparse methods ($p < 0.001$). The results are presented in \tableref{tab:simulation_score}.
They indicate that SPCA-TV is more robust to variation in the learning samples than the other sparse methods. SPCA-TV yields reproducible loading vectors across data sets.

\begin{table}[!t]
    \begin{center}
        \caption{Scores are averaged across the 50 independent data sets. We tested whether
                 the scores obtained with existing PCA methods are
                 significantly different from scores obtained with SPCA-TV. Significance
                 notations: ***:~$p\leq10^{-3}$}
        \begin{tabular}{lccc}
            \toprule
                           & \multicolumn{3}{c}{Scores}\\
            \cmidrule{2-4}
            Methods        & Reconstruction error  & MSE               & Dice Index \\
            \midrule
            Sparse PCA     & 1577.6***           & 0.90***           & 0.28*** \\
            ElasticNet PCA & 1577.0***           & 0.83***           & 0.26*** \\
            \rowcolor{verylightgray}
            SPCA-TV         & 1575.5\phantom{***} & 0.62\phantom{***} & 0.54\phantom{***} \\
            \bottomrule
        \end{tabular}
        \label{tab:simulation_score}
    \end{center}
\end{table}

These results indicate that the loadings are not only more stable across resampling but also achieve a better recovery of the underlying variability in data than the Sparse PCA and ElasticNet PCA methods.

\subsection{3D images of functional MRI of patients with schizophrenia}

\begin{figure}[!bh]
    \centering
    \includegraphics[width=0.4825\textwidth]{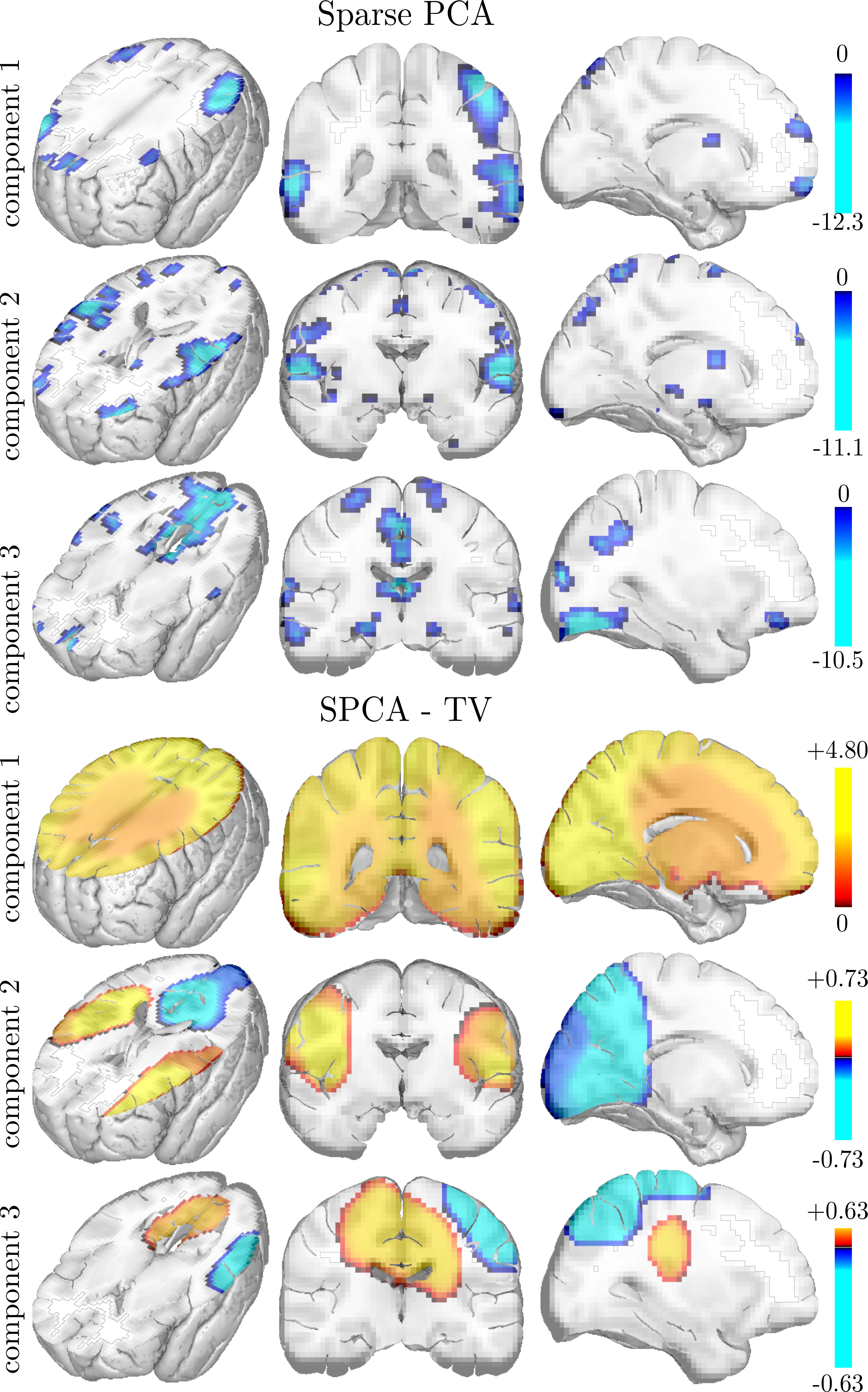}
    \caption{Loading vectors recovered from the 83 activation maps using Sparse PCA and SPCA-TV.}
    \label{fig:Principal_Components_extracted_fmri}
\end{figure}

We then applied the methods on an fMRI data set composed of 23 patients experiencing hallucinations while lying in the scanner. We computed activation maps from the scans preceding hallucinations by regressing for each block the signal time course on a linear ramp function: Indeed, we hypothesized that activation in some regions presents a ramp-like increase during the time preceding the onset of hallucinations. The activation maps that we used as an input to the SPCA-TV method are the statistical parametric maps associated to the coefficients of the block regression. We obtained a data set of $n=83$ maps and  $p = 63\,966$ features. We hypothesized that the principal components extracted with SPCA-TV from these activation maps could uncover major trends of variability within pre-hallucination patterns. Thus, they might reveal the existence of subgroups of patients, according to the sensory modality involved during hallucinations. We decided to retrieve the first three components, accounting for 15\% of the total variance. Since the next PCs offer very little increase in the total explained variance, we stopped retrieving components. The loading vectors extracted from the activation maps of pre-hallucinations scans  with Sparse PCA and SPCA-TV are presented in \figref{fig:Principal_Components_extracted_fmri}. We observe a similar behavior as in the synthetic example, namely that the loading vectors of Sparse PCA tend to be scattered and produce irregular patterns. However, PCA-TV seems to yield structured and smooth sources of variability, which can be interpreted clinically. Furthermore, the SPCA-TV loading vectors are not redundant and revealed different patterns. 

Indeed, the loading vectors obtained by SPCA-TV are of great interest because they revealed insightful patterns of variability in the data: the second loading is composed of interesting areas such as the precuneus cortex and the cingulate gyrus, but also areas related to vision-processing areas such as the occipital fusiform gyrus and the parietal operculum cortex regions. The third loading reveals important weights in the middle temporal gyrus, the parietal operculum cortex and the frontal pole. The first loading vectors is less informative since it appears to encompass all features, but could reveal a variability affecting the whole brain. These results seem to indicate the possible existence of subgroups of patients according to the hallucination modalities involved.  Indeed, we might be able to distinguish patients with visual hallucinations from those suffering mainly from auditory hallucinations by looking at the score of the second component extracted by SPCA-TV.

The reconstruction error is significantly lower in SPCA-TV than in both Sparse PCA ($T=-8.1$, $p=0.0012$) and ElasticNet PCA ($T=-4.7$, $p=0.009$). Moreover, when assessing the stability of the loading vectors across the folds, we found a higher statistically significant mean Dice index in SPCA-TV compared to the Sparse PCA ($p<0.001$) and compared to ElasticNet PCA 's performance ($p=0.008$) as presented in \tableref{tab:fmri_score}.

\begin{table}[!th]
    \begin{center}
        \caption{Scores of the fMRI data are averaged across the 5 folds. We tested
                 whether the averaged scores obtained with existing PCA methods are
                 significantly different from scores obtained with SPCA-TV. Significance
                 notations: ***:~$p\leq10^{-3}$, **:~$p\leq10^{-2}$.}
        \begin{tabular}{lcc}
            \toprule
                     &  \multicolumn{2}{c}{Scores}\\
            \cmidrule{2-3}
            Methods  &  Reconstruction error  & Dice Index \\
            \midrule
            Sparse PCA     & 1548**           & 0.42*** \\
            ElasticNet PCA & 1518**           & 0.57**\phantom{*} \\
            \rowcolor{verylightgray}
            SPCA-TV         & 1474\phantom{**} & 0.70\phantom{***} \\
            \bottomrule
        \end{tabular}
        \label{tab:fmri_score}
    \end{center}
\end{table}

In conclusion, SPCA-TV significantly outperforms Sparse PCA and ElasticNet PCA in terms of the reconstruction error, and in the sense that loading vectors are both more clinically interpretable and more stable.

We also evaluated the convergence speed of Sparse PCA, Mini-Batch Sparse PCA (a variant of Sparse PCA that is faster but less accurate), ElasticNet PCA and SPCA-TV for this functional MRI data set of $n=83$  samples and $p=63\,966$ features. We compared the time of execution required for each algorithm to achieve a given level of precision in \tableref{tab:algos-convergence_mri}. Sparse PCA and ElasticNet PCA are similar in term of convergence time, while mini-batch sparse PCA is much faster but does not converge to high precision. As expected, SPCA-TV takes longer than other sparse methods because of the inclusion of spatial constraints, but the convergence time is still reasonable for an fMRI data set with $65\,000$ voxels.

\begin{table}[!t]
    \begin{center}
        \caption{Comparison of the execution time required for each sparse
                 method to reach the same precision. Times reported in seconds.}
        \begin{tabular}{lrrrrr}
            \toprule
                    &  \multicolumn{5}{c}{Time to reach a given precision in seconds}\\
            \cmidrule{2-6}
            Methods &$10$ & $1$ & $10^{-1}$ & $10^{-2}$ & $10^{-3}$ \\
            \midrule
            Mini-batch Sparse PCA & 5.32 & -  & -  & - & - \\
            Sparse PCA & 158.0 & 231.2 & 344.3 & 386.8 & 450.1   \\
            ElasticNet & 123.7 & 138.1 & 302.7 & 396.4 & 406.3 \\
            \rowcolor{verylightgray}
            SPCA-TV & 427.7 & 2958.6 & 8093.0 & 13813.4 & 14459.9 \\
            \bottomrule
        \end{tabular}
        \label{tab:algos-convergence_mri}
    \end{center}
\end{table}

\subsection{2D meshes of cortical thickness in Alzheimer disease}

Finally, SPCA-TV was applied to the whole brain anatomical MRI from the ADNI database, the Alzheimer's Disease Neuroimaging Initiative, (\url{http://adni.loni.usc.edu/}).
The MR scans are T1-weighted MR images acquired at $1.5~T$ according to the ADNI acquisition protocol. 
We selected 133 patients with a diagnosis of mild cognitive impairments (MCI) from the ADNI database who converted to AD within two years during the follow-up period. We used PCA to reveal patterns of atrophy explaining the variability in this population. This could provide indication of possible stratification of the population into more homogeneous subgroups, that may be clinically similar, but with different brain patterns. In order to demonstrate the relevance of using SPCA-TV to reveal variability in any kind of imaging data, we worked on meshes of cortical thickness.
 The $317\,379$ features are the cortical thickness values at each vertex of the cortical surface. Cortical thickness represents a direct index of atrophy and thus is a potentially powerful candidate to assist in the diagnosis of Alzheimer's disease (\cite{Bakkour2009}, \cite{Dickerson2009}). Therefore, we hypothesized that applying SPCA-TV to the ADNI data set would reveal important sources of variability in cortical thickness measurements. Cortical thickness measures were performed with the FreeSurfer image analysis suite (Massachusetts General Hospital, Boston, MA, USA), which is documented and freely available for download online (\url{http://surfer.nmr.mgh.harvard.edu/}). The technical details of this procedure are described in \cite{Sled1998}, \cite{Dale1999} and  \cite{Fischl1999a}. All the cortical thickness maps were registered onto the FreeSurfer common template (fsaverage). We decided to retrieve the first three components, accounting for 8\% of the total variance. Since the next PCs offer very little increase in the total explained variance, we stopped retrieving components.

\begin{figure}[!bh]
    \centering
    \includegraphics[width=0.4825\textwidth]{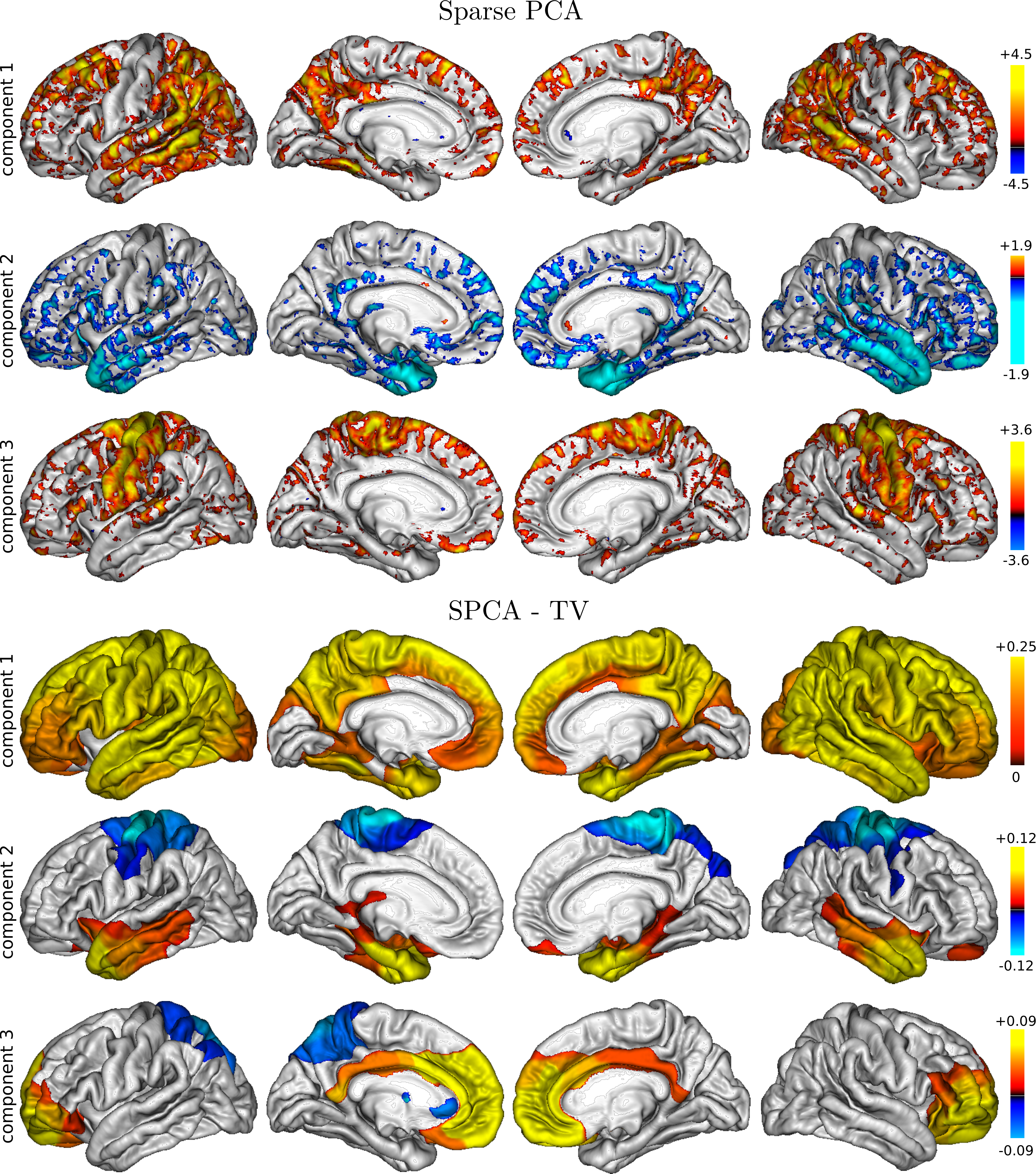}
    \caption{Loading vectors recovered from the 133 MCI patients using Sparse PCA and SPCA-TV.}
    \label{fig:Principal_Components_extracted_adni}
\end{figure}

The loading vectors obtained from the data set with sparse PCA and SPCA-TV are presented in \figref{fig:Principal_Components_extracted_adni}. As expected, Sparse PCA loadings are not easily interpretable because the patterns are irregular and dispersed throughout the brain surface. On the contrary, SPCA-TV reveals structured  and smooth clusters in relevant regions. The first loading vector, which maps the whole surface of the brain, can be interpreted as the variability between controls and  Alzheimer converters, resulting from a global brain atrophy. The second loading vector includes variability in the entorhinal cortex, hippocampus and in temporal regions. And last, the third loading vector might be related to the atrophy of the frontal lobe and also involved variability in the precuneus. Thus, SPCA-TV provides a smooth map that closely matches the well-known brain regions involved in Alzheimer's disease.

Indeed, it is well-documented that cortical atrophy progresses over three main stages in Alzheimer disease. The cortical structures are sequentially being affected because of the accumulation of amyloid plaques. Cortical atrophy is first observed, in the mild stage of the disease, in regions surrounding the hippocampus, as seen in the second component. Then, the disease progresses to a moderate stage; where atrophy gradually extends to the prefrontal association cortex as revealed in the third component. In the severe stage of the disease, the whole cortex is affected (first component). Therefore, SPCA-TV provides us with clear biomarkers, that are perfectly relevant in the scope of Alzheimer's disease progression.

The reconstruction error is significantly lower in SPCA-TV than in both Sparse PCA ($T=-35.2$, $p=4\cdot10^{-6}$) and ElasticNet PCA ($T=-9.9$, $p=5\cdot10^{-4}$). The results are presented in \tableref{tab:Comparison_reconstruction_stability_ADNI}.
Moreover, when assessing the stability of the loading vectors across the folds, the mean Dice index is significantly higher in SPCA-TV than in both Sparse PCA ($p=0.003$) and ElasticNet PCA ($p<0.004$).

\begin{table}[!ht]
    \begin{center}
        \caption{Scores are averaged across the 5 folds. We tested whether the averaged
                 scores obtained with existing PCA methods are significantly lower from scores obtained with SPCA-TV. Significance notations:
                 ***:~$p\leq10^{-3}$, **:~$p\leq10^{-2}$.}
        \begin{tabular}{lcc}
            \toprule
                           &  \multicolumn{2}{c}{Scores} \\
            \cmidrule{2-3}
            Methods        &  Reconstruction error  & Dice Index \\
            \midrule
            Sparse PCA     &  2863***              & 0.57** \\
            ElasticNet PCA &  2844***              & 0.64** \\
            \rowcolor{verylightgray}
            SPCA-TV         &  2817\phantom{***}  & 0.78\phantom{**} \\
            \bottomrule
        \end{tabular}
        \label{tab:Comparison_reconstruction_stability_ADNI}
    \end{center}
\end{table}

\section{Conclusion}

We proposed an extension of Sparse PCA that takes into account the spatial structure of the data. The optimization scheme is able to minimize any combination of the $\ell_1$, $\ell_2$, and TV penalties while preserving the exact $\ell_1$ penalty. We observe that SPCA-TV, in contrast to Sparse PCA and ElasticNet PCA, yields clinically interpretable results and reveals major sources of variability in data, by highlighting structured clusters of interest in the loading vectors. Furthermore, SPCA-TV 's loading vectors were more stable across the learning samples compared to Sparse PCA. SPCA-TV was validated and its applicability was demonstrated on three distinct data sets: we may reach the conclusion that SPCA-TV can be used on diverse data, and is able to present structure within the data.  

\section*{Supplementary Material}

\paragraph{The ParsimonY Python library}
\begin{itemize}
\item\textit{Url:} \url{https://github.com/neurospin/pylearn-parsimony}
\item\textit{Description:} ParsimonY is Python library for structured and sparse machine learning.
 ParsimonY is open-source (BSD License) and compliant with scikit-learn API.
\end{itemize}

\paragraph{Data sets and scripts}
\begin{itemize}
\item\textit{Url:}  \url{ftp: //ftp.cea.fr//pub/unati/brainomics/papers/pcatv}
\item\textit{Description:} This url provides the simulation data set and the Python script used to create Fig.2 for the paper.
\end{itemize}


\bibliographystyle{plain}
\bibliography{pca_tv}

\end{document}